\documentclass[hidelinks]{article}

\usepackage[preprint]{neurips_2021}

\usepackage{natbib} 

\usepackage{mathtools} 
\usepackage{booktabs} 
\usepackage{todonotes}

\usepackage[utf8]{inputenc} 
\usepackage[T1]{fontenc}    
\usepackage{hyperref}       
\usepackage{url}            
\usepackage{caption} 
\usepackage{amsfonts}       
\usepackage{nicefrac}       
\usepackage{microtype}      
\usepackage{xcolor}         
\usepackage{siunitx}        

\newcommand\Tstrut{\rule{0pt}{2.ex}}         
\newcommand\Bstrut{\rule[-1.ex]{0pt}{0pt}}   
\newcommand\TBstrut{\Tstrut\Bstrut}           
\usepackage{hyperref}

\usepackage[ruled,vlined]{algorithm2e}
\newlength\mylen

\SetKwInput{KwInput}{Input} 
\SetKwInput{KwParams}{Hyper-parameters}

\usepackage[american]{babel}
\usepackage[mathscr]{euscript}

\usepackage{mathtools} 
\usepackage{booktabs} 
\usepackage{tikz} 
\usepackage{amssymb,bm}
\usepackage[ruled,vlined]{algorithm2e}
\usepackage{subcaption}

\title{WeatherFusionNet: Predicting Precipitation from Satellite Data 
}

\author{%
  Jiří Pihrt\\
    Faculty of Information Technology\\
    Czech Technical University in Prague\\
  \texttt{pihrtjir@fit.cvut.cz} \\
\And
  Rudolf Raevskiy\\
    Faculty of Information Technology\\
    Czech Technical University in Prague\\
  \texttt{raevsrud@fit.cvut.cz} \\
  \And
    Petr Šimánek\\
    Faculty of Information Technology\\
    Czech Technical University in Prague\\
  \texttt{petr.simanek@fit.cvut.cz} \\
  \And
    Matej Choma\\
    Meteopress spol s.r.o.\\
    Faculty of Information Technology\\
    Czech Technical University in Prague\\
  \texttt{matej.choma@meteopress.cz} \\
}

\begin{document}

\maketitle

\begin{abstract}
    The short-term prediction of precipitation is critical in many areas of life. Recently, a large body of work was devoted to forecasting radar reflectivity images. The radar images are available only in areas with ground weather radars. Thus, we aim to predict high-resolution precipitation from lower-resolution satellite radiance images. A neural network called WeatherFusionNet is employed to predict severe rain up to eight hours in advance. WeatherFusionNet is a U-Net architecture that fuses three different ways to process the satellite data; predicting future satellite frames, extracting rain information from the current frames, and using the input sequence directly. Using the presented method, we achieved 1st place in the NeurIPS 2022 Weather4cast Core challenge. The code and trained parameters are available at \url{https://github.com/Datalab-FIT-CTU/weather4cast-2022}.
\end{abstract}

\section{Introduction}\label{sec:intro}
Reliably predicting precipitation is a challenge with societal impact, especially the prediction of extreme weather. The Weather4Cast 2022 \citep{iarai} competition allowed us to work on the nowcasting problem for areas with no or unreliable access to weather radar data. Using only satellite data to forecast extreme precipitation is a very challenging task, and it complements our present effort to master radar prediction. Weather4Cast 2022 follows the previous challenge of  \cite{weather4cast}.

The joint team of Meteopress and Czech Technical University in Prague aims to develop the best possible approach for precipitation prediction. We want to overcome the limitations of the current optical methods by applying machine learning. We currently work on improving nowcasting based on weather radar data with very accurate results as presented by \cite{choma2022precipitation}. In the past, we used a basic U-Net \cite{unet} and later PredRNN architecture \citep{predrnn} to solve this challenging task. We further improved the models, but we later found it necessary to include more prior physical knowledge in the architecture. The PhyDNet of \cite{phydnet} demonstrated superior performance in radar prediction (in both storm structure and movement) while increasing the interpretability of the model and also the physical trustworthiness. We further improved the physical part of PhyDNet \citep{choma2022precipitation}. 

In the Weather4Cast, the task is even more challenging than radar prediction. We decided to apply our good experience with PhyDNet and enrich the model with other architectural choices and fusion. We call the resulting model WeatherFusionNet as it fuses three different ways to process the satellite data; predicting future satellite images, extracting rain information from the current frames, and using the input sequence directly.

\section{Used data}
The task of this competition was to predict 32 ground-radar (1 channel) images, which is responsible for the next 8 hours, based on 4 satellite images (1 hour). The entire dataset was provided by the organizers of the Weather4cast competition. It covers areas of 7 European regions and contains satellite and radar images for years 2019 and 2020. The satellite images are composed of 2 visible (VIS), 2 water vapor (WV), and 7 infrared (IR) bands. Each image has a size of $252 \times 252$ and has a temporal resolution of 15 minutes. Satellite patch pixel corresponds to 12x12 square kilometers. The target rain rate in the sequence to be predicted have higher spatial resolution and 1 pixel corresponds to a spatial area of about 2x2 square kilometers. 

Training sequences are generated with a sliding window, using the provided data loader, generating a total of 228928 samples. The validation set contains predefined sequences, 840 in total.

Static data with latitude, longitude, and topological height were also provided, but were not used in our model.

The competition was designed as a binary classification task. The rain threshold was specified as 0.2 during the second stage.

\section{Model Architecture}\label{sec:arch}

Our model consists of three modules, which are all trained separately, looking at the data from different angles.

\begin{figure}[h]
  \centering\leavevmode
  \includegraphics[width=1.\linewidth]{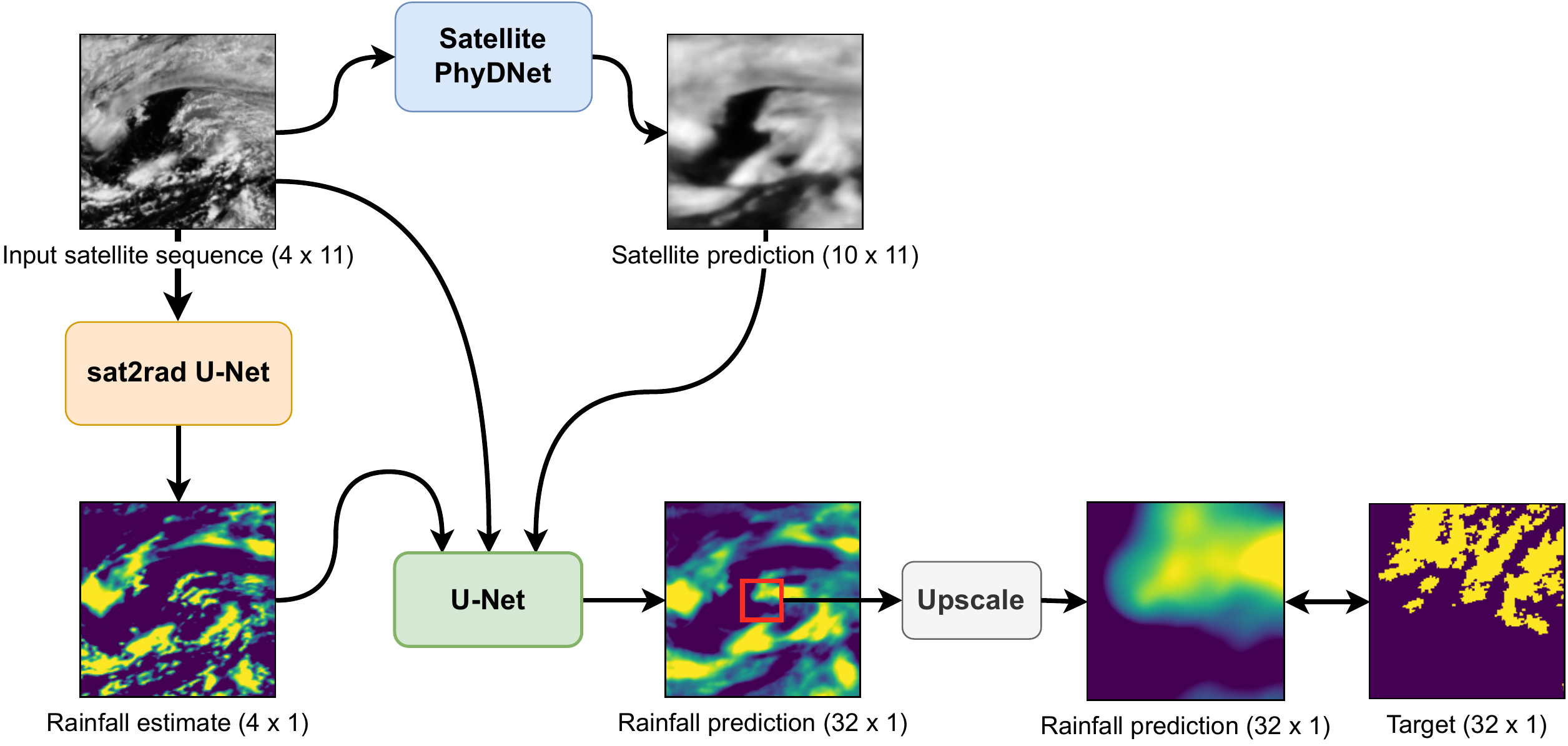}
  \caption{Diagram of the model architecture. The dimensions in parentheses denote the temporal and channel dimension sizes, respectively.}
  \label{fig:model-diagram}
\end{figure}

Firstly, we use a module that we call \textbf{sat2rad}. This network is trained to estimate the rainfall at the current time step of a single satellite frame. By training it this way, we believe it can efficiently extract information about the current rain situation in the input sequence, without having to predict the future. This module is realized by a U-Net~\citep{unet} with 11 input channels and 1 output channel. In the overall model, it is applied to all 4 input satellite input frames independently, generating 4 channels in total. We take advantage of the spatial invariance feature of a convolutional network, to predict the rainfall for the entire satellite area, even though we only have radar data for a smaller area. More details on that, as well as the specific U-Net architecture, are described later in this section.

Secondly, we use a recurrent convolutional network PhyDNet~\citep{phydnet}. PhyDNet aims to disentangle physical dynamics from other complementary visual information. Its architecture consists of two branches. One branch is responsible for the physical dynamics, and features a recurrent physically constrained cell called PhyCell, for performing PDE-constrained prediction in latent space. The second branch extracts other residual information required for future prediction, such as visual appearance and details, using ConvLSTM cells \citep{shi2015convolutional}.

Although PhyDNet is designed to work with radar frames, it was difficult to apply it to this competition in a straightforward way, because of the combination of relatively small spatial area and long prediction timeframe, and the fact that we don't  have accurate input radar data during inference. Instead, we trained PhyDNet only on the satellite data, which do not have these limitations. Essentially, we used PhyDNet to extend the input sequence of satellite frames. We decided to limit the output sequence length to 10, based on our memory limitations and past experience with PhyDNet.

Finally, the outputs from the previously described two modules are concatenated, along with the input sequence, and fused by another U-Net to generate the final prediction. This U-Net has a total of 158 input channels and 32 output channels. However, similarly, as in the sat2rad module, the prediction covers the entire $3024\times3024$ square kilometer area. But we only need to predict and have the data for, the center $504\times504$ square kilometer area. Because of that, we crop the center part ($42\times42$ pixels) of the prediction, and then upscale it back to the target $252\times252$ resolution, as shown in Figure~\ref{fig:model-diagram}. We believe this approach can take more advantage of the spatial invariance feature of a convolutional network such as U-Net, than simply forcing it to cover the smaller spatial area directly. We also use this approach when training the sat2rad module.

The upscale operation is realized simply by an Upsample layer with bilinear interpolation and a scale factor of 6. We tried more sophisticated approaches during the first stage of the competition, but we did not observe any significant improvements. There is room for more research on this part.

For both of the U-Net modules in our architecture, we use the following version;
\begin{itemize}
    \item filter sizes of 32, 64, 128, 256, 512,
    \item each 3x3 convolution is followed by a BatchNorm layer and a ReLU,
    \item downscaling is realized by 2x2 max pooling with stride 2
    \item and upscaling is realized by Upsample layers with a scale factor of 2 and bilinear interpolation.
\end{itemize}
One limitation of U-Net is that it only works with images of sizes divisible by 32. To get around this, we pad the input images by 2 pixels (with padding mode set to replicate) on each side to increase the size to 256x256, right before inputting them to both U-Nets we use. Then we crop the output back to 252x252.

\section{Training Procedure}\label{sec:train}
As described in the previous section, the sat2rad module was not trained on sequences, but on individual frames. To achieve this, we simply set the input and output sequence length to 1 and subtracted 1 from the output indexes so that they match the input. No further modifications were required. It was trained with a batch size of 32, and the loss function used was BCEWithLogitsLoss. Other hyperparameters are listed in Table~\ref{tab:hyper}.

\begin{table}[h!]
    \centering
    \begin{tabular}{|l|l|}
        \hline
        Learning rate                              & 1e-3   \Tstrut  \\ \hline
        Weight decay                   & 1e-2   \Tstrut  \\ \hline
        Optimizer                       & AdamW   \Tstrut \\ \hline
        Model precision                       & 32  \Tstrut     \\ \hline
        Positive weight & 2.58 \TBstrut \\ \hline
    \end{tabular}
    \caption{U-Net hyperparameters}
    \label{tab:hyper}
\end{table}

The Satellite PhyDNet module is trained only using satellite data. We set the satellite sequence length to 14 and then split it into 4 input and 10 output frames. Because the provided validation set was not designed for longer sequences, we used part of the training set as a validation split. Specifically, the first 150000 samples were used for training, and the rest was used to generate validation sequences, using a sliding window with a stride of 32 samples. PhyDNet was trained with a loss function $L=L1+L2$. We used teacher forcing, starting with probability of 1, decreased by 5e-5 every step. Hyperparameters are listed in Table \ref{tab:phydnet}. 

\begin{table}[h!]
    \centering
    \begin{tabular}{|l|l|}
        \hline
        \textbf{ConvLSTMCell}                   &  \Tstrut   \\ \hline
        \quad Input dimension &  64 \Tstrut \\ \hline
        \quad Hidden dimensions & [128, 128, 64] \Tstrut\\ \hline
        \quad Kernel size & (3, 3)  \Tstrut   \\ \hline
        \textbf{PhyCell}                   &    \Tstrut  \\ \hline
        \quad Input dimension &  64\Tstrut \\ \hline
        \quad Hidden dimensions & [49]\Tstrut \\ \hline
        \quad Kernel size & (7, 7)    \Tstrut \\ \hline
        Optimizer                       & Adam \Tstrut   \\ \hline
        Learning rate                       & 1e-3 \Tstrut   \\ \hline
        Batch Size & 16\TBstrut \\ \hline
    \end{tabular}
    \caption{PhyDNet hyperparameters}
    \label{tab:phydnet}
\end{table}

\begin{figure}[h!]
      \centering\leavevmode
      \includegraphics[width=1.\linewidth]{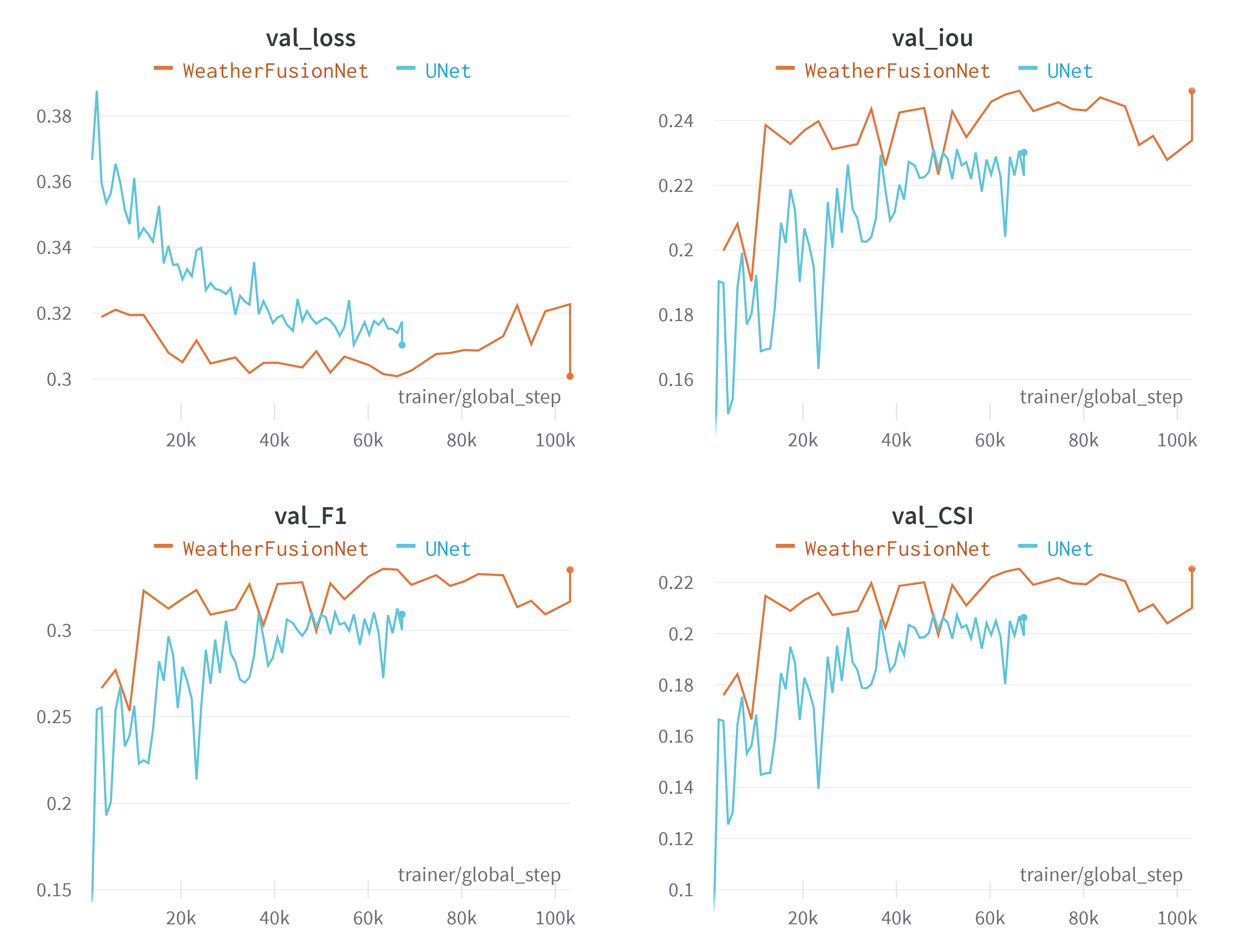}
      \caption{Evolution of the validation loss, IoU, F1, and CSI metric during training. Compared WeatherFusionNet with plain UNet.}
      \label{fig:val_loss}
    \end{figure}

The final U-Net model requires outputs from the previous two modules, but during training they were frozen (their weights were not updated). This model was also trained with BCEWithLogitsLoss, batch size 16, and other hyperparameters listed in Table~\ref{tab:hyper}. No learning rate scheduling was used. The evolution of CSI, F1, IoU metrics and loss on the validation dataset is presented in Figure \ref{fig:val_loss}. The models have been trained using a single NVIDIA A100-40GB.

\section{Results}\label{sec:results}

    In this section, we show the empirical results of our model. We also compare it to a baseline U-Net model, which has exactly the same hyperparameters and training procedures, but it does not include the inputs from sat2rad and PhyDNet modules. It is difficult to explain why the WeatherFusionNet performed worse on the Transfer IoU without access to the test and heldout dataset.
    
    \begin{table}[h!]
        \centering
        \begin{tabular}{ c||c|c|c } 
        \textbf{Model} & \textbf{Validation IoU} & \textbf{Core Heldout IoU} & \textbf{Transfer Heldout IoU} \Bstrut \\ 
        \hline
        UNet & 0.2988 & 0.2950 & \fontseries{b}\selectfont 0.2567 \Tstrut \\
        WeatherFusionNet & \fontseries{b}\selectfont 0.3212  & \fontseries{b}\selectfont 0.3162 & 0.2488
        \end{tabular}
        \caption{IoU metric on the validation set and from the final competition leaderboards.}
        \label{tab:results}
    \end{table}
    
    \begin{figure}[h!]
      \centering\leavevmode
      \includegraphics[width=1\linewidth]{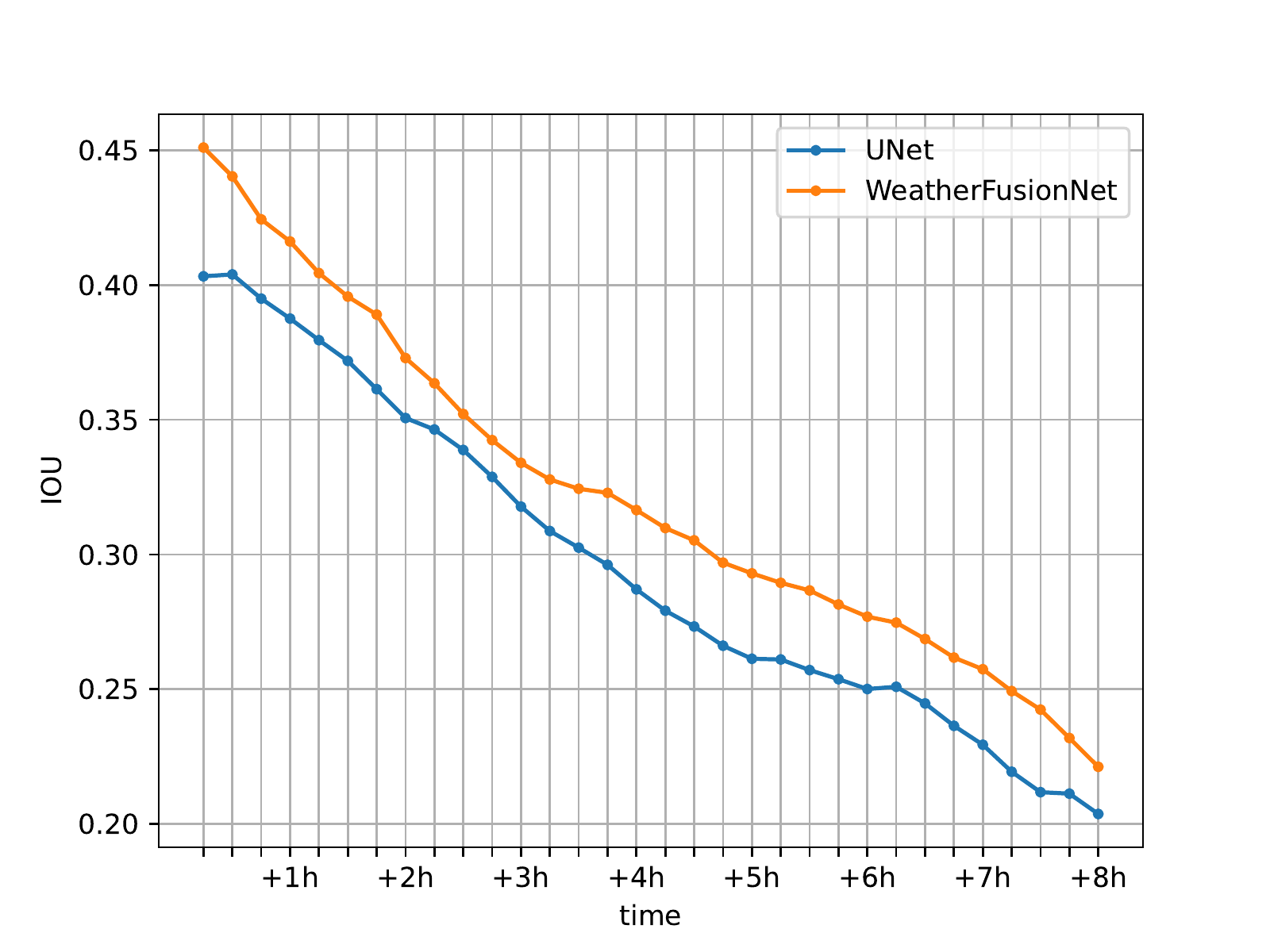}
      \caption{IoU metric over time computed on the validation set.}
      \label{fig:iou-over-time}
    \end{figure}
    
\begin{figure}[h!]
      \centering\leavevmode
      \includegraphics[width=1.\linewidth]{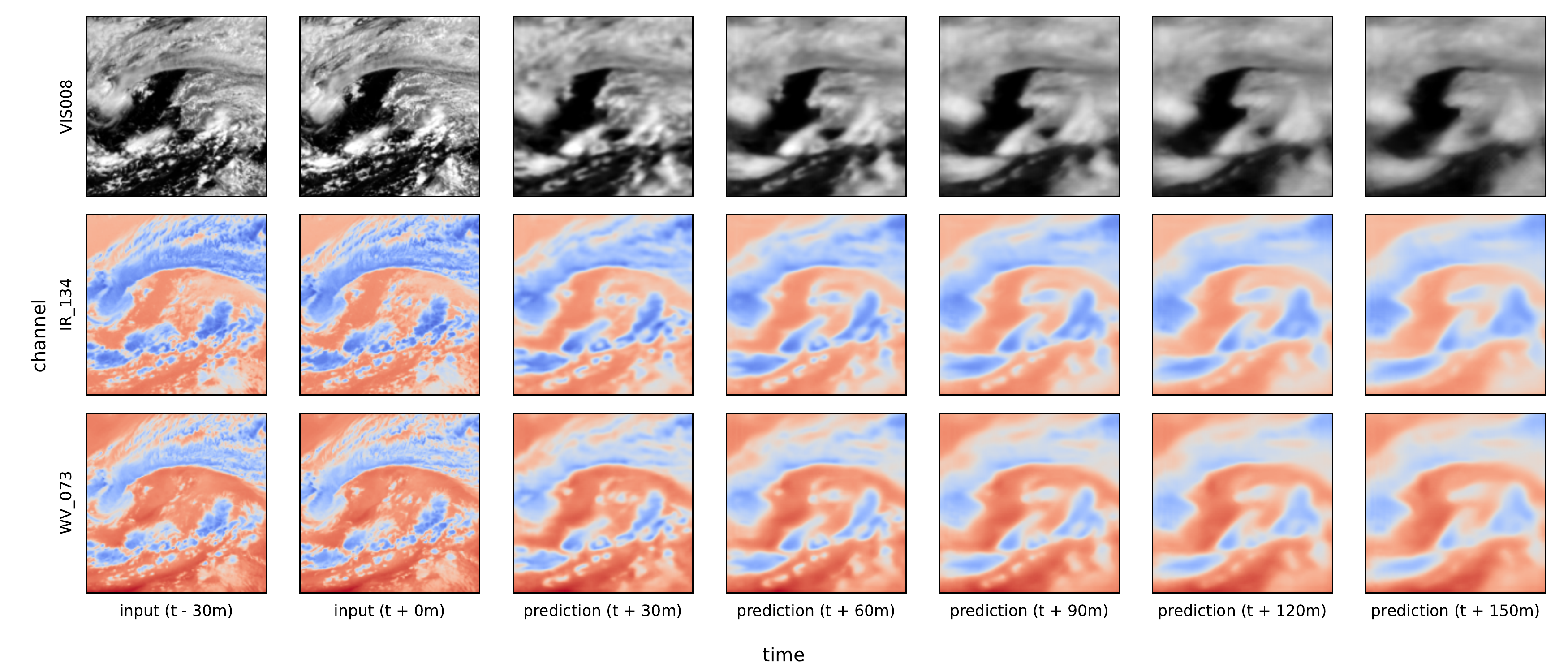}
      \caption{Satellite PhyDNet prediction example. Each row is a different satellite channel, first two columns are input data, and further columns show PhyDNet prediction of the satellite for up to 150 minutes.}
      \label{fig:phydnet-out}
    \end{figure}

\begin{figure}[h!]
      \centering\leavevmode
      \includegraphics[width=0.7\linewidth]{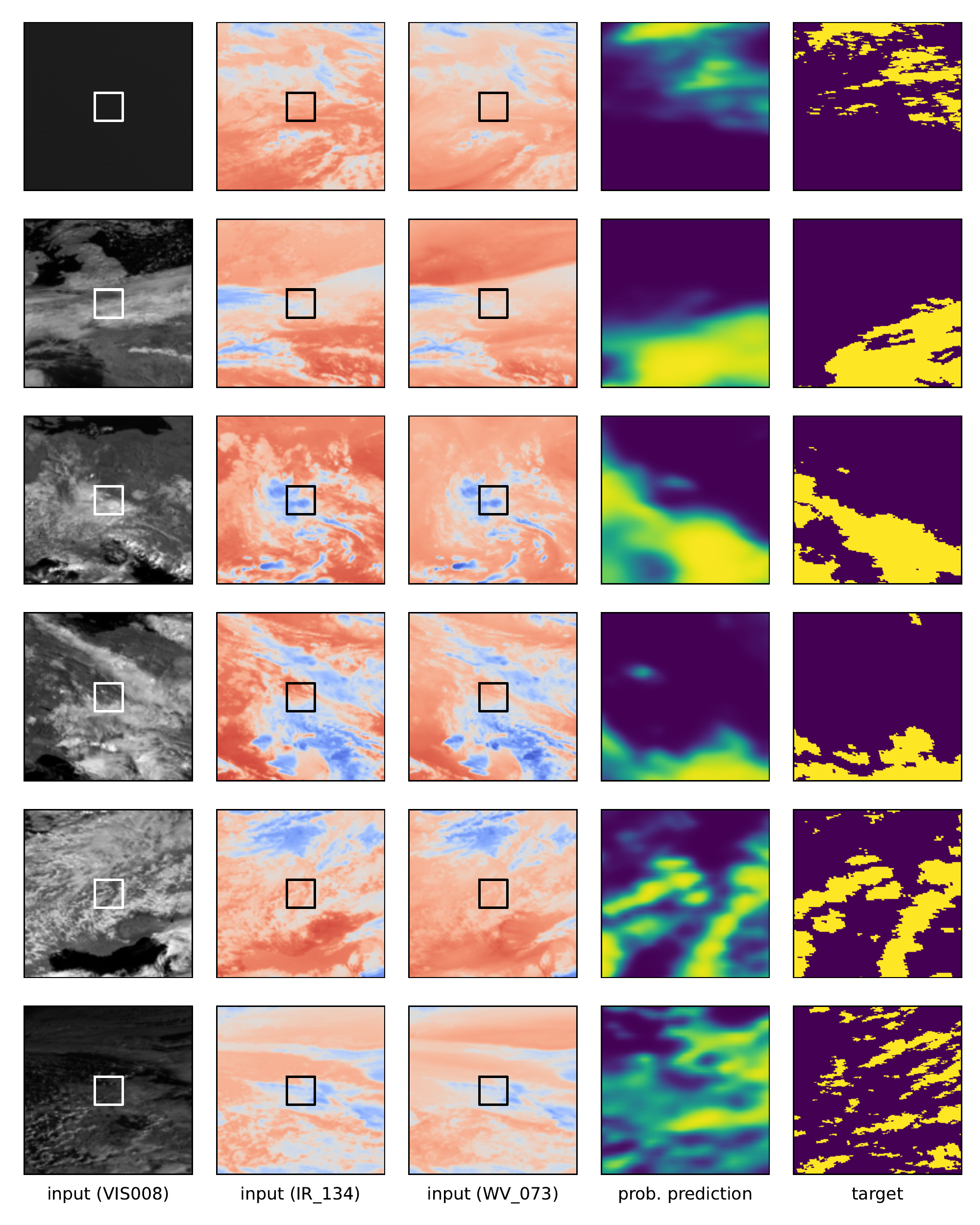}
      \caption{sat2rad U-Net module prediction examples. Each row illustrates one sample time instance, first three columns show input satellite data (three different channels). The black/white square highlights the target radar area. The fourth column presents predicted rain probability and the last column is the target radar image.}
      \label{fig:sa2rad-out}
    \end{figure}

\begin{figure}[h!]
      \centering\leavevmode
      \includegraphics[width=1.\linewidth]{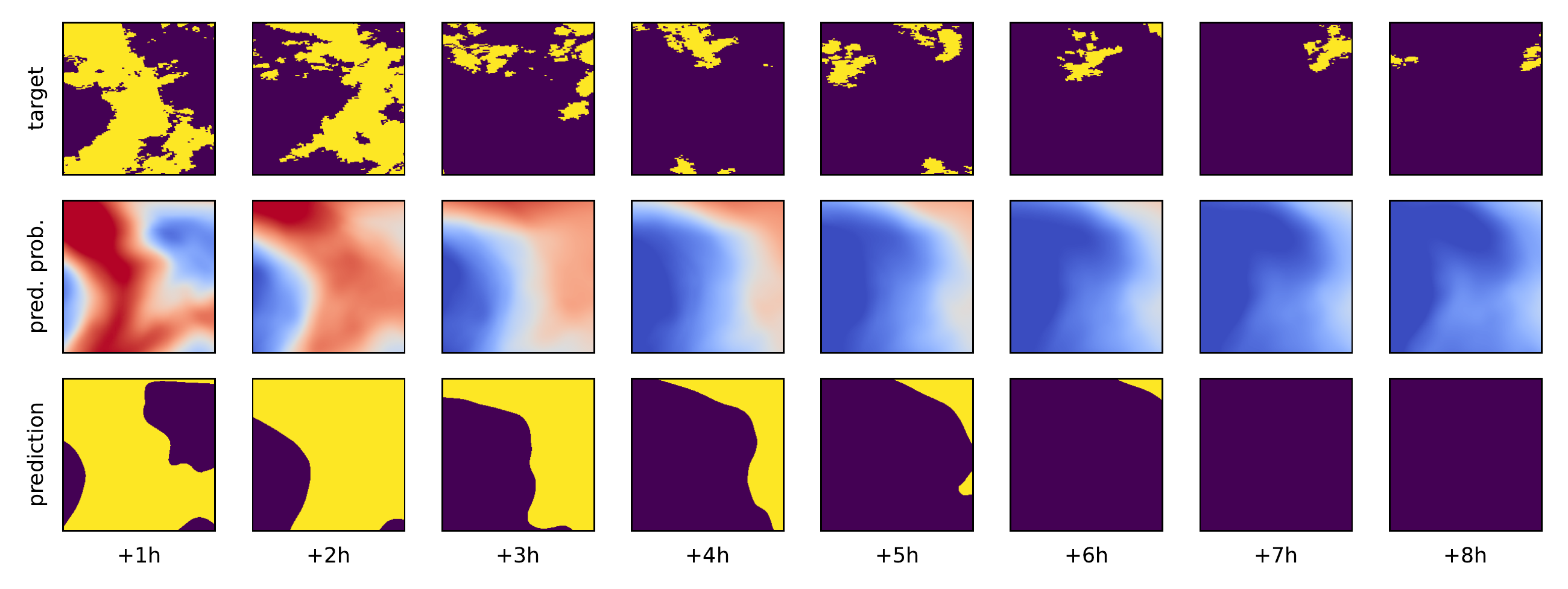}
      \caption{WeatherFusionNet prediction example. The first row is the evolution of the target radar image in 8 hours, the second row is the predicted probability and the last row is the final prediction.}
      \label{fig:predictions}
    \end{figure}

    Figure~\ref{fig:iou-over-time} shows how the IoU metric varies in time over a prediction. Figure~\ref{fig:phydnet-out} demonstrates a~prediction of a satellite image sequence by PhyDNet. Figure~\ref{fig:sa2rad-out} shows results of sat2rad U-Net module. Figure~\ref{fig:predictions} presents a sample WeatherFusionNet prediction. 

\section{Conclusion}\label{sec:conclusion}

We presented our approach to forecasting precipitation in high resolution based only on low-resolution satellite data. The method is called WeatherFusionNet and was used as a solution to the Weather4cast challenge. The model ingests three different sources of data, i.e. from satellite prediction, sat2rad, and the satellite image. The model proved to be working well by winning the Weather4cast 2022 Core challenge.

The current model is not trained end-to-end, mostly due to memory requirements, but it would be interesting to try it in the future. Special attention should be also paid to the upscaling part of the model. The upscaling is not critical in the current setting but would be if we would try to solve a regression task instead of classification. Especially to model the storm with a reasonable structure which is a very difficult task for the current deep-learning methods. Also, we may add static data that has not been used in this paper.

The model is also well prepared for another source of data, especially if the radar data are available, we can skip the sat2rad model.

\clearpage

\bibliography{template.bib}

\end{document}